%% file: main.tex
\documentclass[10pt,twocolumn,letterpaper]{article}

\usepackage[pagenumbers]{cvpr}

\usepackage{graphicx}
\usepackage{amsmath}
\usepackage{amssymb}
\usepackage{booktabs}
\usepackage{comment}
\usepackage{stfloats}

\usepackage[pagebackref,breaklinks,colorlinks]{hyperref}
\usepackage{placeins}

\usepackage[capitalize]{cleveref}
\crefname{section}{Sec.}{Secs.}
\Crefname{section}{Section}{Sections}
\Crefname{table}{Table}{Tables}
\crefname{table}{Tab.}{Tabs.}

\begin{document}

\definecolor{darkgreen}{rgb}{0, 0.6, 0.3}
\definecolor{lightgray}{gray}{0.6}

\newcommand{\OURS}{ROCA}

\title{\OURS{}: Robust CAD Model Retrieval and Alignment from a Single Image}

\author{Can G{\"u}meli \\
\and
Angela Dai\\ \\
Technical University of Munich\\
\and
Matthias Nie{\ss}ner \\
}
\title{\OURS{}: Robust CAD Model Retrieval and Alignment from a Single Image}

\author{Can G{\"u}meli 
\qquad\qquad
Angela Dai
\qquad\qquad
Matthias Nie{\ss}ner \\
\\
Technical University of Munich 
}

% Teaser
\begin{figure}
\twocolumn[{%
\renewcommand\twocolumn[1][]{#1}%
\maketitle
\begin{center}
    \centering
    %\vspace{-10mm}
    \includegraphics[width=0.95\textwidth]{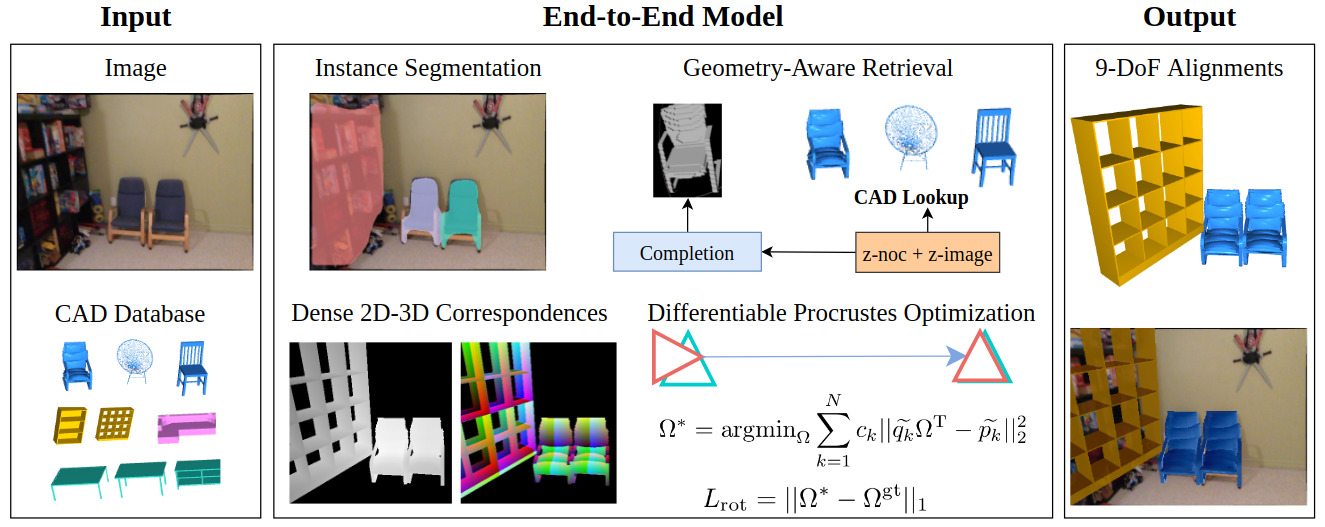}
    \vspace{-3mm}
    \caption{We present \OURS{}, a new end-to-end model that robustly retrieves and aligns 3D CAD models to a single image. From an RGB image and a database of CAD models, \OURS{} retrieves and aligns CAD models for each object in the image. In comparison to previous methods that perform direct pose regression, our approach leverages differentiable Procrustes optimization by predicting dense 2D-3D correspondences in the form of depths and normalized object coordinates (NOCs). In addition, predicted 3D correspondences help to learn retrieval of geometrically similar CAD models while simultaneously improving object alignments.}
    \label{figure:teaser}
    \vspace{-1mm}
\end{center}%
}]
\end{figure}

\maketitle

\input{c0_abstract}

\input{c1_intro}

\input{c2_related_work}

\begin{figure*}
    \centering
    \includegraphics[width=\textwidth]{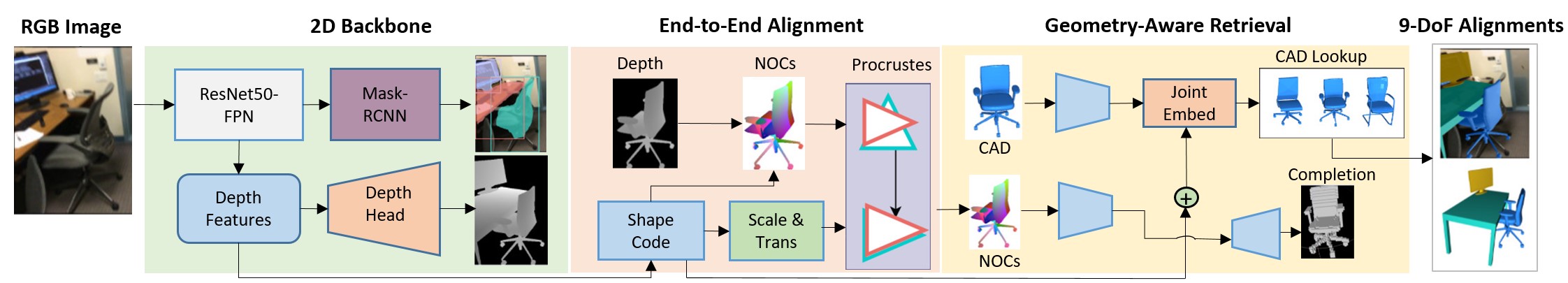}
    \vspace{-0.7cm}
    \caption{\textbf{Method Overview.} 
    From an input RGB image, we first estimate 2D instance segmentation and dense depth for each object.
    For each detected object, we then estimate dense correspondences to its canonical space as NOCs, which inform our differentiable Procrustes optimization for geometrically-informed object alignment.
    We additionally leverage the NOCs to construct a joint embedding space between estimated NOCs and voxelized CAD models to enable CAD retrieval, leveraging a proxy geometric completion loss.
    This enables geometrically informed CAD retrieval, and moreover, robust CAD alignment to the image.
    }
    \label{figure:model}
\end{figure*}

\input{c3_method}
\input{c4_results}

\input{c5_conclusion}
\input{c6_acknowledge}

{\small
\bibliographystyle{ieee_fullname}
\bibliography{main}
}

\appendix

\input{c7_appendix}

\end{document}

%% file: c0_abstract.tex
\begin{abstract}
We present \OURS{}\footnote{Code and project video are available through our project page, \url{https://cangumeli.github.io/ROCA/}.}, a novel end-to-end approach that retrieves and aligns 3D CAD models from a shape database to a single input image.
This enables 3D perception of an observed scene from a 2D RGB observation, characterized as a lightweight, compact, clean CAD representation.
Core to our approach is our differentiable alignment optimization based on dense 2D-3D object correspondences and Procrustes alignment.
\OURS{} can thus provide a robust CAD alignment while simultaneously informing CAD retrieval by leveraging the 2D-3D correspondences to learn geometrically similar CAD models.
Experiments on challenging, real-world imagery from ScanNet show that \OURS{} significantly improves on state of the art, from 9.5\% to 17.6\% in retrieval-aware CAD alignment accuracy.
\end{abstract}

%% file: c1_intro.tex
\section{Introduction}

2D perception systems have seen remarkable advances in object recognition from 2D images in recent years, enabling widespread adoption of systems that can perform accurate 2D object localization, classification, and segmentation from an image~\cite{imagenet, fpn, maskrcnn}.
Such advances have spurred forward developments in many fields, from classical image understanding to robotics and autonomous driving. 
However, unlike the human perception of 2D images, these systems tend to perform object recognition purely in 2D, whereas from a single RGB image, a human can perceive geometric shape, structure, and pose of the objects in the scene.
In fact, such 3D understanding is crucial for many applications, enabling possible exploration and interaction with an observed environment.

At the same time, there has been notable progress in estimating 3D object geometry from visual data~\cite{pixel2mesh, occupancy, ogn, psgn, atlasnet, 3dr2n2}. 
In particular, Mesh R-CNN~\cite{meshrcnn} introduced a formative approach to 3D object estimation from real-world images, bridging state-of-the-art 2D object detection with voxel-to-mesh estimation of the shape of each detected object in an image.
In contrast to this generative approach, Mask2CAD proposed to retrieve and align CAD models from a database to produce a lightweight object reconstruction with high fidelity given by the CAD database~\cite{mask2cad, vid2cad}.
With significant availability of synthetic CAD models~\cite{modelnet, shapenet, 3dfuture}, such CAD model reconstruction of an observed scene shows strong promise in perceiving 3D from an image, as it enables geometric estimation in a clean, compact fashion more akin to artist-crafted 3D models, and easily consumed for downstream applications.

However, current approaches to estimate 3D object structure from an RGB image have largely focused on shape representation and generation, either without any explicit pose estimation~\cite{meshrcnn}, or simply regressing the object pose directly from the 2D features of the object~\cite{mask2cad, vid2cad, patch2cad, total3d, points2objects}.
Thus, we propose a new CAD retrieval and alignment approach to estimate 3D perception from an image by formulating a differentiable 9-DoF pose optimization directly coupled to the object retrieval; this enables a more robust, geometry-aware, end-to-end CAD alignment to the image.

In this work, we propose \OURS, a new method that jointly detects object regions in a given input image while simultaneously estimating depth and dense correspondences between each 2D object region and its location in its canonical object space. 
From the dense depth and correspondence estimates, we then formulate a differentiable Procrustes optimization and produce a final set of retrieved CAD models and their 9-DoF alignments.
This geometry-aware differentiable optimization for the CAD alignment of each object enables more robust and accurate CAD alignment. In addition, our method learns geometry-aware embeddings for CAD retrieval. Our retrieval embedding module utilizes the canonical coordinates used in pose optimization and an auxiliary shape completion objective. We show learning geometry-aware embeddings improves both retrieval and alignment accuracy. 
Overall, our method significantly outperforms state of the art, improving by 8.1\% and 9.5\% in retrieval-aware alignment accuracy and alignment accuracy from a single RGB image.

In summary, we propose an end-to-end architecture for CAD model alignment to an RGB image with:
\begin{itemize}
	\item a new differentiable pose optimization enabling geometry-aware 2D-to-3D CAD alignment to an RGB image,
	\item improved CAD retrieval by leveraging the dense object pose correspondences and proxy CAD completion objective to inform the construction of a joint embedding space between detected objects and CAD models,
	\item an interactive runtime of 53 milliseconds per image, facilitating its use in real-time applications.
\end{itemize}

%% file: c2_related_work.tex
\section{Related Work}

\noindent\textbf{2D Object Recognition}. 
Single-image 3D understanding requires strong 2D recognition capabilities. 
With recent advances in deep learning, we now have methods that have achieved remarkable success in image classification \cite{imagenet, resnet}, 2D object detection \cite{fpn, retinanet}, and 2D instance segmentation \cite{maskrcnn, shapemask}. 
We build our approach from the success of 2D recognition to extend to an end-to-end 3D object reasoning.
In particular, to extract image features, we use a Mask-RCNN \cite{maskrcnn} recognition backbone that detects and segments objects in the image, from which we then estimate 3D CAD alignment and retrieval.

\medskip\noindent\textbf{Monocular Depth Estimation}. 
As we aim to predict 9-DoF object alignments, we must reason about absolute object depths.
Monocular depth estimation from a single RGB image has been extensively studied to predict absolute per-pixel depths from an image. Modern monocular depth estimation frameworks typically train from pre-trained deep convolutional networks to predict per-pixel depth prediction in a fully convolutional fashion \cite{eigen}, leveraging learned semantic features to help resolve scale-depth ambiguities. 
Recent methods have proposed using deeper networks, multi-scale feature extractors, and alternative loss functions \cite{ddp, revisiting}. 
To extract strong geometric features, we also predict dense depth for each detected object, leveraging state-of-the-art depth estimation techniques \cite{ddp, revisiting}.

\medskip\noindent\textbf{Single-Image Shape Reconstruction}. 
In recent years, many deep learning-based approaches have been developed to reconstruct 3D shapes from a 2D image. 
These approaches typically employ large synthetic 3D shape datasets \cite{shapenet}, and render the synthetic objects for image input.
Many different shape representations have been explored, including voxel grids \cite{3dr2n2, ogn}, point clouds \cite{psgn}, polygonal meshes \cite{pixel2mesh, atlasnet, scan2mesh}, and neural implicit functions \cite{deepsdf,occupancy}.
Such approaches have seen significant success in the single-object scenario, inspiring Mesh-RCNN \cite{meshrcnn} to pioneer an approach built on 2D object recognition that generates 3D shapes for each detected object in an RGB image, as a first step to 3D perception in real-world images. 
More recent work has also focused on estimating scene layouts and inter-object relations \cite{relnet, total3d}.
In contrast to these approaches that perform object-based reconstruction from an image, we propose to optimize for object poses by establishing dense geometric correspondences, rather than a direct regression, enabling more robust alignment optimization.

\medskip\noindent\textbf{CAD Model Retrieval and Alignment}. 
3D reconstruction using CAD model priors has a long history in computer vision \cite{chin1986model,roberts1963machine,binford1982survey}. 
Recently, with the availability of large-scale 3D shape datasets \cite{shapenet}, several approaches have been introduced to perform CAD model retrieval and alignment to an image based on analysis-by-synthesis reconstruction \cite{im2cad, holistic}. 
While promising, these methods have extensive computational costs (minutes to hours), making them ill-suited for many potential image-based reconstruction applications, such as real-time mobile or robotic applications.
More recently, Mask2CAD \cite{mask2cad} proposed a more lightweight approach that learns to simultaneously retrieve and 5-DoF align 3D CAD models to detected objects in an image, building on top of a state-of-the-art 2D recognition backbone \cite{retinanet, shapemask}.
Its use of a CAD representation for lightweight object-based reconstruction and perception has inspired a patch-based approach to improve CAD retrieval \cite{patch2cad}, as well as extension to video input in Vid2CAD with full 9-DoF alignment for each object (estimating scale and depth, nor originally predicted by Mask2CAD) \cite{vid2cad}.
We also aim to align and retrieve CAD models to reconstruct the objects in an RGB image, but rather than directly regressing the object pose, we formulate a differentiable optimization based on learned, dense object correspondences, enabling more accurate CAD alignment.
Our established dense object correspondences also enable explicit geometry-aware retrieval, as we establish a joint embedding space between 3D CAD and 3D object correspondences.

Several recent approaches also solve for 9-DoF object alignment in an image. For instance, Points2Objects \cite{points2objects} builds from 2D object detection \cite{centernet} to directly regress 9-DoF alignments and treats object retrieval as a classification problem, primarily demonstrating results on synthetic renderings.
In contrast to their direct bounding box regression for alignment,  we formulate a differentiable alignment based on dense geometric correspondence and perform nearest neighbor object retrieval from a CAD database.

\medskip\noindent\textbf{Learned Differentiable Pose Optimization}. Traditional pose optimization leverages correspondence finding to solve for pose alignment, with separate stages for establishing correspondences and the optimization for the best pose under the correspondences.
With the introduction of deep learning, several methods have been proposed to bridge such pose optimization into an end-to-end learned pipeline, by establishing learned correspondences to inform various pose optimization tasks, such as point cloud registration \cite{registration}, PnP optimization \cite{pnp}, or non-rigid tracking \cite{nonrigid}. 
Most inspirational to our work is the differentiable Procrustes optimization of \cite{e2esocs} for  CAD model alignment to RGB-D scans. 
In contrast to their work, we must reason without 3D input data, and thus formulate a notably different optimization to focus on image-based object pose estimation with per-pixel correspondences and a weighted formulation for robust alignment estimation.

%% file: c3_method.tex
\section{Method}

\subsection{Overview}
From an input RGB image $I$, its camera intrinsics $\pi$, and a database of 3D CAD models $\mathcal{S}$, we aim to represent each object in the image as a similar CAD model in its 9-DoF alignment to the image in metric camera space, thus providing a comprehensive lightweight, geometric reconstruction of the observed scene.
An overview of our approach is shown in Figure~\ref{figure:model}.

From the RGB image, we first detect and segment objects in 2D with a Mask-RCNN \cite{maskrcnn} backbone, while simultaneously estimating the object depths with a multi-scale FPN \cite{fpn}. 
We then estimate the 9-DoF alignment for a detected object by using its estimated depths, 2D features, and instance mask to regress scale and an initial translation.
Moreover, we also establish learned, dense correspondences to the object's canonical space.
We then formulate a  weighted Procrustes optimization to solve for the rotation and refined translation, simultaneously estimating the optimization weights for robust alignment optimization. 
Our approach is trained end-to-end with supervision given by CAD models that have been aligned to RGB images.

In addition to alignment, our method learns to retrieve geometrically similar CAD models to represent each detected object.
We first use the predicted Mask-RCNN object categories to determine the object class from which to search. 
We establish geometric similarity by learning geometry-aware joint embeddings between the detected objects and CAD models, leveraging the  detected objects' canonical correspondences and 2D features; this helps to significantly bridge the domain gap of 2D observations and 3D CADs.
To additionally learn shape-guided features, our embedding spaces are also trained jointly with a proxy shape completion objective. 
This retrieval is trained end-to-end along with the object alignment predictions and enables CAD retrieval as a nearest-neighbor lookup from the input CAD database at inference time.

\subsection{Object Recognition and Depth Estimation}
From an input image $I$, we leverage a strong 2D backbone to estimate both 2D object recognition and dense depth.
We use a ResNet-50-FPN \cite{resnet, fpn} to form the 2D backbone, which produces a feature map $F$ that then informs both the 2D instance segmentation following Mask-RCNN \cite{maskrcnn}, as well as the depth estimation using a Multi-scale Feature Fusion (MFF) \cite{revisiting} module.
The 2D object recognition detects objects as a set of object bounding boxes $\{b_i\}$ and instance masks $\{m_i\}$, and the depth estimation produces a features map $F^d$ from the MFF layer before outputting the estimated depth map $D$. 
For 2D object recognition, we use a Mask-RCNN \cite{maskrcnn} backbone, and for depth estimation, we use a ResNet-50-FPN \cite{resnet, fpn} with a Multi-scale Feature Fusion \cite{revisiting} module that up-projects \cite{ddp} and concatenates four levels of FPN features.
To up-sample depth to full image resolution, we use a pixel-shuffle layer \cite{superres}.
We use a masked reverse Huber (berHu) \cite{ddp} loss to optimize for depth map $D$. 

We additionally refine the predicted depths in the regions of detected objects.
For each detected object $i$ with predicted 2D bounding box $b_i$, we consider the $d_i$ cropped from $D$ using $b_i$ and resized with nearest-neighbor interpolation to an ROI size of $32\times 32$.
We then apply an additional loss to focus on object depth estimation, $||d_i - d_i^{\mathrm{gt}}||_1 + ||\mu(d_i) - \mu(d_i^{\mathrm{gt}})||_1,$
where $\mu$ represents the average function.

We can then project the estimated per-object depths $d_i$ with the intrinsics $\pi$ to obtain camera-space positions $p_i$.

\subsection{Robust Differentiable CAD Alignment} 
\label{section:alignment}

For each detected object, we then estimate its 9-DoF alignment to the image as its translation $\mathbf{t}_i$, scale $\mathbf{s}_i$, and rotation $\mathbf{r}_i$.

\noindent\textbf{Aggregating region features}.
To estimate the alignment for a detected object $i$,with predicted 2D bounding box $b_i$ and instance mask $m_i$, we first aggregate its corresponding features from $F^d$ which encode strong geometric descriptors of the object.
We crop the features from $F^d$ using ROIAlign~\cite{maskrcnn} in the box region $b_i$, resulting in $f_i^d$ of dimension $32\times 32\times 128$.
For each object region, we then use a robust feature aggregation scheme to obtain a shape descriptor $\mathbf{e}_i$:
 \begin{equation}
    \mathbf{e}_i = \textrm{MLP}(\textrm{MaxPool}(m_i \odot \textrm{PixelMLP}(f_i^d)))
 \end{equation}
where $\odot$ denotes the element-wise product operator, and $\textrm{PixelMLP}$ is a fully-connected network that operates per-pixel and is shared over the pixels.
Both MLPs use two layers with hidden size 512, and $\textrm{MaxPool}$ performs a max pooling over the pixels in the object region to a single feature, to help increase robustness against mask and detection errors. The final $\textrm{MLP}$ output activation contains a ReLU as well as a 30\% dropout regularization.

\medskip\noindent\textbf{Scale regression}.
We directly use predicted shape codes to regress scale, $\mathbf{s}_i = \textrm{MLP}(\mathbf{e}_i)$, and optimize it with an $\ell_1$ loss:
\begin{equation}
    L_{\mathrm{scale}} = ||\mathbf{s}_i - \mathbf{s}_i^{\mathrm{gt}}||_1.
\end{equation}
Note that the final affine layer produces scale estimates for each class category, from which the scale is selected based on the class category prediction, enabling capture of any class-specific scale characteristics.

\medskip\noindent\textbf{Initial translation estimation}.
We additionally predict an initial translation estimate, based on the back-projected predicted object depths, $p_i$, as a learned offset $\mathbf{t}^o_i$ from the center of the $p_i$: 
\begin{align*}
    \mathbf{t}^{\mathrm{init}}_i=0.5(\max(p_i)+\min(p_i)) + \mathbf{t}^o_i,\\
    \mathbf{t}^o_i = \mathrm{MLP}([\mathbf{e}_i, (\max(p_i)-\min(p_i)])).
\end{align*}
where $[\cdot,\cdot]$ denotes concatenation, and the MLP contains a single hidden layer of size 256.

We then apply losses on the $p_i$ and $\mathbf{t}^{\mathrm{init}}_i$:
\begin{equation}
    L_p = \mathrm{H}\left([\max(p_i), \min(p_i)], [\max(p_i^{\mathrm{gt}}), \min(p_i^{\mathrm{gt}})]\right)
\end{equation}
\begin{equation}
    L_{\mathrm{trans\_initial}} = ||\mathbf{t}_i^{\mathrm{init}} - \mathbf{t}_i^{\mathrm{gt}}||_2,
\end{equation}
where $\mathrm{H}$ denotes the Huber loss.
Similar to the scale regression, we output per-category estimates from which the final output is selected based on the predicted class category.

\medskip\noindent\textbf{Normalized Object Coordinates as Correspondences}.
In order to predict the final translation $\mathbf{t}_i$ and rotation $\mathbf{r}_i$ of the object, we propose to predict dense correspondences to the canonical object space, characterized as normalized object coordinates, to inform the alignment optimization. 
In contrast to a direct regression, this enables the alignment prediction to be geometrically informed and to learn the correspondences that best inform the alignment energy.

That is, for each foreground pixel in $m_i$, we predict its corresponding 3D coordinates in the normalized object coordinate (NOC) space, where the object lies in its canonical orientation is normalized into a unit cube $[-0.5, 0.5]^3$. Since NOCs will represent the same geometric surfaces as the back-projected depths, we predict NOC correspondences $q_i$ based on the inverse translated depth points, predicted scale, and shape code, using a lightweight MLP with two 256-unit hidden layers:
\begin{equation}
    q_i = \textrm{MLP}([p_{i} - \mathbf{t}_i^{\mathrm{init}}, \mathbf{s}_i, \mathbf{e}_i]).
\end{equation}
We optimize for predicted NOCs using an $\ell_1$ loss w.r.t. ground-truth NOCs:
\begin{equation}
    L_{\mathrm{noc}} = ||q_i - q_i^{\mathrm{gt}}||_1.
\end{equation}

\medskip\noindent\textbf{Differentiable Robust Procrustes Optimization}. 
At the core of our alignment prediction, we formulate a differentiable weighted Procrustes optimizer that estimates rotation and refined translation. 
For each object, we have the regressed scale and translation $\mathbf{s} \in \mathbb{R}^{3}$ and $\mathbf{t}^{\mathrm{init}} \in \mathbb{R}^{3}$, NOC-depth correspondences $(q, p) \in \mathbb{R}^{\textrm{N} x 3}$, and $m$ contains the segmentation mask probability of each foreground NOC-depth correspondence. 
For notational simplicity, now we omit the index suffix $i$ from the previous sections.

Our objective is to solve for an orthogonal rotation $\Omega \in \mathbb{SO}_3$ such that $(q \odot \mathbf{s})\Omega^\mathrm{T} = (p - \mathbf{t})$. We formulate this problem as a least squares optimization:
\begin{equation}
    \Omega^{*} = \mathrm{argmin}_{\Omega} \sum_{k=1}^N c_k ||\widetilde{q_k}\Omega^\mathrm{T} - \widetilde{p_k}||_2^2,
\end{equation}
where $\tilde{q} = q \odot \mathbf{s}$, $\tilde{p} = p - \mathbf{t}$, and $c \in \mathbb{R}^{\textrm{N}}$ is a set of learned, positive weights that allow for learning to upweight more reliable correspondences. 
We predict $c$ using a network conditioned on shape code $\mathbf{e}$ and network predictions: $c_k = \textrm{MLP}([\mathbf{e}, \mathbf{s}, m_k, q_k, \widetilde{p_k}])$. 
This weight prediction MLP contains a single 256-unit hidden layer followed  by a sigmoid. 
The optimization problem is then solved analytically by differentiable weighted Procrustes optimization~\cite{registration}:
\begin{equation}
    \mathrm{U} \Sigma \mathrm{V} = \textrm{SVD}(\tilde{p}^{\textrm{T}}
    \begin{bmatrix}c_{1} & & \\ & \ddots & \\
    & & c_{\textrm{N}}
  \end{bmatrix}
 \tilde{q})
\end{equation}
\begin{equation}
    \Omega^{*} = \mathrm{U} \begin{bmatrix} 1 &  &  \\  & 1 &  \\  &  & \textrm{det}(\mathrm{V} \mathrm{U}^\textrm{T})\end{bmatrix} \mathrm{V}^\textrm{T}
\end{equation}
We optimize for the resulting rotation with an $\ell_1$ loss using the ground-truth rotation matrix:
\begin{equation}
    L_{\mathrm{rot}} = ||\Omega^{*} - \Omega^{\mathrm{gt}}||_1.
\end{equation}
Procrustes optimization can also be used to estimate translation by first zero-centering points for rotation and then using the resulting rotation and point averages for translation~\cite{registration}.
We found that directly estimating translation this way resulted in unstable training due to large transformations. To this end, we instead solve for a refined translation $\mathbf{t}$ based on the initial estimate $\mathbf{t}^{\mathrm{init}}$:
\begin{equation}
    \mathbf{t} = \mathbf{t}^{\mathrm{init}} + \mu_{c}(\mathrm{\widetilde{p}}) - \Omega^{*} \mu_{c}(\mathrm{\widetilde{q}}),
\end{equation}
where $\mu_{\mathbf{c}}$ denotes the weighted average operation w.r.t. predicted optimization weights $c$.
We optimize for the refined translation with the  objective:
\begin{equation}
    L_{\mathrm{trans}} = ||\mathbf{t} - \mathbf{t}^{\mathrm{gt}}||_2.
\end{equation}

Thus, our final alignment loss is:
\begin{equation*}
    L_{\mathrm{align}} = w_{\mathrm{rot}}L_{\mathrm{rot}} + L_{\mathrm{trans}} + L_{\mathrm{scale}} + w_{\mathrm{noc}}L_{\mathrm{noc}} + L_{\mathrm{trans\_initial}}.
\end{equation*}
We can then train for alignment in an end-to-end fashion, directly informing the predicted geometric correspondences learned from the image input, resulting in more robust alignment estimates.

\subsection{Geometry-Aware CAD Retrieval} \label{section:retrieval}
In addition to geometric object alignment, we propose a geometry-aware image-to-CAD retrieval that learns a joint embedding space of CAD models and object regions.

To represent a detected object $i$ in the image, we use its shape code and predicted NOCs described in Section~\ref{section:alignment}:
\begin{equation}
    \mathbf{z}_i = \textrm{MLP}(\mathbf{e}_i) + \textrm{3DCNN}(\textrm{Voxelize}(q_i)),
\end{equation}
where $\mathbf{z}_i\in \mathbb{R}^{256}$, determined by the combination of an $\textrm{MLP}$ that processes the shape code $\mathbf{e}_i$ with a single hidden unit of size 1024 and a final ReLU activation, as well as a lightweight 3D convolutional network $\textrm{3DCNN}$ that operates on the $32^3$ voxelized NOCs (the architecture is adopted from \cite{jointemb}, we refer to the supplemental for more architecture details).
To differentiably voxelize the NOCs, we use the trilinear point cloud voxelization from PyTorch3D~\cite{pytorch3d}, and normalize the resulting density grid such that each occupancy probability is $\in [0, 1]$.

To encode the CAD models, we similarly represent them as $32^3$ occupancy grids $G$, which are then encoded with a 3D CNN:
\begin{equation}
    \mathbf{\tilde{z}} = \textrm{3DCNN}(G).
\end{equation}
Here, the $\textrm{3DCNN}$ is structured symmetrically to the NOC processing network, but they do not share weights, since NOC and CAD domains remain quite different.

Using the image and CAD representations above, our model learns a joint embedding space. 
For an image region embedding $\mathbf{z}_i$, its corresponding positive and negative CAD embeddings $\mathbf{\tilde{z}}_i^{+}$ and $\mathbf{\tilde{z}}_i^{-}$, we construct the space with a triplet loss:
\begin{equation}
    L_{\mathrm{ret}} = \max(||\mathbf{z}_i - \mathbf{\tilde{z}}_i^{+}||_2^2 - ||\mathbf{z}_i - \mathbf{\tilde{z}}_i^{-}||_2^2 + 0.5, 0).
\end{equation}

Since our predicted NOCs represent only the visible object geometry rather than the complete object geometry, whereas CAD models represent complete objects, we use an additional proxy completion loss on the NOC encoding:
\begin{equation}
    L_{\mathrm{comp}} = \textrm{BCE}(\textrm{3DUpCNN}(\mathbf{z}_i), \mathrm{G}_i^{+})
\end{equation}
where $\mathrm{BCE}$ is the binary cross-entropy loss, $\mathrm{G}_i^{+}$ is the positive CAD's occupancy grid, and $\textrm{3DUpCNN}$ a 3D convolutional decoder that predicts the complete object geometry from the NOC encoding.

Our geometry-aware retrieval is trained end-to-end with the alignment prediction, allowing for retrieval to also inform the predicted NOC correspondences as well as the image features.
At inference time, we pre-compute CAD embeddings and perform a nearest-neighbor lookup for each detected object region embedding prediction.

\subsection{Implementation Details}
\noindent\textbf{Training}.  
We train our approach end-to-end, using a momentum optimizer with momentum 0.9 and an  initial learning rate of 1e-3, starting with 1k warm-up iterations interpolating from 1e-4 \cite{maskrcnn}. 
The learning rate is decayed by 0.1 at iteration 60k, and training takes 80k iterations for convergence.
We use a batch size of 4, and 128 region proposals. 
To regularize our model, we use a weight decay of 1e-4 and apply random contrast, brightness, and saturation augmentation, all sampled from the intensity range $[0.8, 1.25]$. 
We use loss weights $w_{\mathrm{rot}}=2, w_{\mathrm{noc}}=3$ to balance them with the translation and scale losses. 
To address class imbalances, we weight detection, segmentation, and alignment losses inversely proportional to the logarithm of training class frequencies. 
Note that we do not weight retrieval losses, as more frequent classes need better retrieval. 
Our 2D recognition backbone is pre-trained on ImageNet and then COCO \cite{imagenet, coco}.
In total, training takes  $\approx 20$ hours on a single RTX 3090 GPU.

\medskip\noindent\textbf{Inference}. 
For 2D object recognition at inference time, we follow Mask-RCNN \cite{maskrcnn}, filtering detections using non-maxima suppression.
Object detections with $<0.5$ confidence are discarded, and we use a mask probability threshold of $0.7$ to estimate foreground object regions for more robust feature aggregation and alignment.
For retrieval, we use a $0.5$ threshold for region segmentation.
Although our model is not optimized for speed, running the full inference pipeline only takes 53 ms ($\approx$ 19 frames per second), resulting in near real-time, interactive running speed.

\medskip\noindent\textbf{Implementation}. Our model implementation is based on PyTorch \cite{pytorch}, Detectron2 \cite{detectron2}, and PyTorch3D \cite{pytorch3d} frameworks. We use the weights of Detectron2 Mask-RCNN-R50-FPN-1x model to initialize our recognition backbone.

%% file: c4_results.tex
\begin{table*}[ht]
\centering
 \resizebox{0.9\textwidth}{!}{
\begin{tabular}{
p{0.17\linewidth}|
p{0.05\linewidth}
p{0.04\linewidth}
p{0.04\linewidth}
p{0.05\linewidth}
p{0.05\linewidth}
p{0.05\linewidth}
p{0.05\linewidth}
p{0.04\linewidth}
p{0.04\linewidth}
|p{0.07\linewidth} p{0.07\linewidth}
}

Method & bathtub & bed & bin & bkshlf & cabinet & chair & display & sofa & table & class & instance \\
\hline 
Total3D-ODN~\cite{total3d} & 10.0 & 2.9 & 16.8 & 2.8 & 4.2 & 14.4 & 13.1 & 5.3 & 6.7 & 8.5 & 10.4 \\
MDR-CN~\cite{mdr} & 5.8 & 5.7 & 0.9 & 9.9 & 5.4 & 28.1 & 11.5 & 11.5 & 8.1 & 9.7 & 15.3 \\
Mask2CAD-b5~\cite{mask2cad} & 8.3 & 2.9 & 25.9 & 3.8 & 5.4 & 30.9 & 17.3 & 5.3 & 7.1 & 11.9 & 17.9 \\
\hline
Ours P & 14.2 & 7.1 & 18.5 & 6.1 & 13.1 & 30.3 & 17.8 & \textbf{17.7} & 9.2  & 14.9 \tiny{\textcolor{lightgray}{(+3.0)}} & 19.3 \tiny{\textcolor{lightgray}{(+1.4)}}\\

Ours P+E & 21.7 & 4.3 & 27.2 & 13.7 & 15.0 & \textbf{42.9} & 22.0 & 13.3 & 14.5  & 19.4 \tiny{\textcolor{lightgray}{(+7.5)}} & 26.9 \tiny{\textcolor{lightgray}{(+9.0)}}\\

Ours P+E+W  & 18.3 & 10.0 & 28.9 & \textbf{15.1} & 14.6 & 41.4 & 23.0 & 15.9 & \textbf{16.1}  & 20.4 \tiny{\textcolor{lightgray}{(+8.5)}} & 27.0 \tiny{\textcolor{lightgray}{(+9.1)}} \\

\hline
Ours P+E+W+R & \textbf{22.5} & \textbf{12.9} & \textbf{29.7} & 11.3 & 13.8 & 40.6 & 28.3 & 12.4 & 14.8 & 20.7 \tiny{\textcolor{lightgray}{(+8.8)}} & 26.7 \tiny{\textcolor{lightgray}{(+8.8)}} \\ 

{\bf Ours} (final) & \textbf{22.5} & 10.0 & 29.3 & 14.2 &  \textbf{15.8} & 41.0 & \textbf{30.4} & 15.9 & 14.6  & \textbf{21.5} \tiny{\textcolor{darkgreen}{(+9.6)}}  & \textbf{27.4} \tiny{\textcolor{darkgreen}{(+9.5)}}\\

\hline

\end{tabular}}
\vspace{-0.2cm}
\caption{
\textbf{Alignment Accuracy on ScanNet \cite{scannet, scan2cad}} in comparison to the state of the art and ablations.
Total3D-ODN and MDR-CN are the 3D object detectors of Total3D \cite{total3d} and MDR-CenterNet \cite{mdr}, respectively. In both detectors, we provide ground-truth rotations in lieu of layout estimation.  Mask2CAD-b5 is Mask2CAD that predicts full 9-DoF alignment \cite{mask2cad,vid2cad}.
In our ablations, P denotes Procrustes alignment optimization, E end-to-end training, W learned weighted optimization, and R learned retrieval without any completion proxy.
Both end-to-end training and weighted robust optimization improve alignment accuracy.
Our learned retrieval and completion additionally improve the alignment performance.
}
\label{table:alignment}
\vspace{-0.2cm}
\end{table*}

\begin{table*}[ht]
\centering
 \resizebox{0.9\textwidth}{!}{\begin{tabular}{
p{0.17\linewidth}|
p{0.05\linewidth}
p{0.04\linewidth}
p{0.04\linewidth}
p{0.05\linewidth}
p{0.05\linewidth}
p{0.05\linewidth}
p{0.05\linewidth}
p{0.04\linewidth}
p{0.04\linewidth}
| p{0.07\linewidth} p{0.07\linewidth}}

Method & bathtub & bed & bin & bkshlf & cabinet & chair & display & sofa & table & class & instance \\
\hline
Mask2CAD-b5~\cite{mask2cad} & 7.5 & 2.9 & 23.3 & 2.8 & 4.2 & 23.0 & 12.0 & 3.5 & 6.0 & 9.5 & 13.8 \\
\hline
Ours P+E+W & 14.2 & 10.0 & 25.4 & \textbf{10.8} & 9.2 & 27.3 & 17.3 & \textbf{15.9} & \textbf{13.0}  & 15.9 \tiny{\textcolor{lightgray}{(+6.4)}} & 19.4 \tiny{\textcolor{lightgray}{(+5.6)}} \\
Ours P+E+W+R & \textbf{21.7} & \textbf{12.9} & \textbf{27.6} & 5.7 & 10.8 & 30.6 & 19.9 & 9.7 & 11.2  & 16.7 \tiny{\textcolor{lightgray}{(+7.2)}} & 20.5 \tiny{\textcolor{lightgray}{(+6.7)}} \\
{\bf Ours} (final) & 20.8 & 10.0 & 26.7 & 8.5 & \textbf{11.9} & \textbf{32.1} & \textbf{22.5} & 14.2 & 11.8 & \textbf{17.6} \tiny{\textcolor{darkgreen}{(+8.1)}} & \textbf{21.7} \tiny{\textcolor{darkgreen}{(+7.9)}} \\

\hline
\end{tabular}}
\vspace{-0.2cm}
\caption{\textbf{Retrieval-Aware Alignment Accuracy on ScanNet \cite{scannet, scan2cad}}. 
Mask2CAD-b5 is Mask2CAD that predicts full 9-DoF alignment \cite{mask2cad,vid2cad}.
We additionally evaluate different retrieval strategies: P+E+W does not use learned retrieval and instead uses Chamfer distance as a retrieval metric, and P+E+W+R introduces learned retrieval but without proxy completion. Our final approach with learned retrieval with proxy completion achieves the best retrieval-aware alignment performance.}
\label{table:retrieval_alignment}
\vspace{-0.3cm}
\end{table*}

\section{Results}
\begin{figure}
\begin{center}
    \centering
    \includegraphics[width=.43\textwidth]{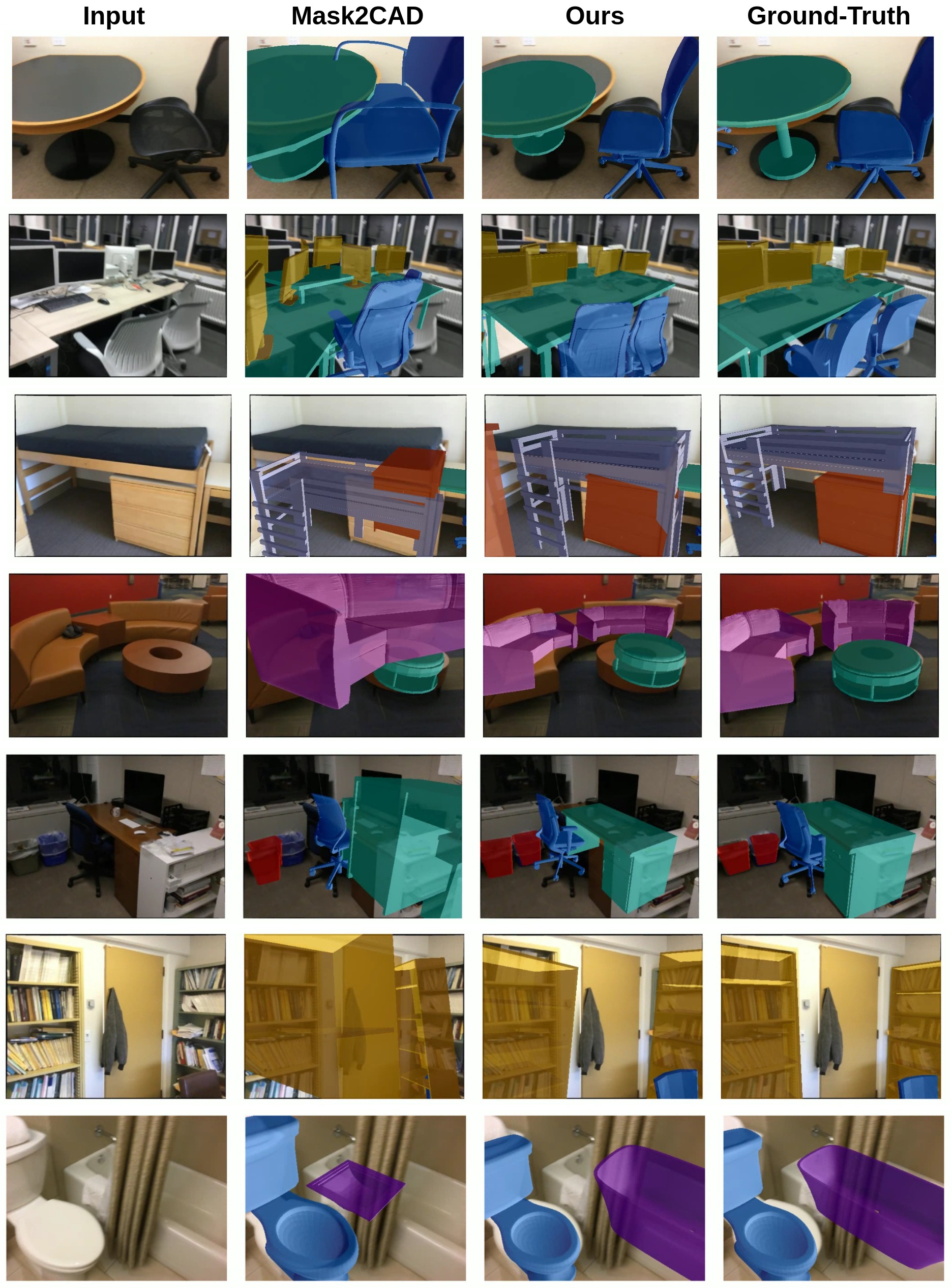}
    \vspace{-3mm}
    \caption{Qualitative evaluation on RGB images from ScanNet \cite{scannet} with Scan2CAD \cite{scan2cad} annotations. 
    \OURS{} obtains more robust and accurate object alignments in many complex scenes.}
    \label{figure:qualitative_comp}
\end{center}
\vspace{-0.95cm}
\end{figure}

\begin{figure*}
    \centering
    \includegraphics[width=0.95\textwidth]{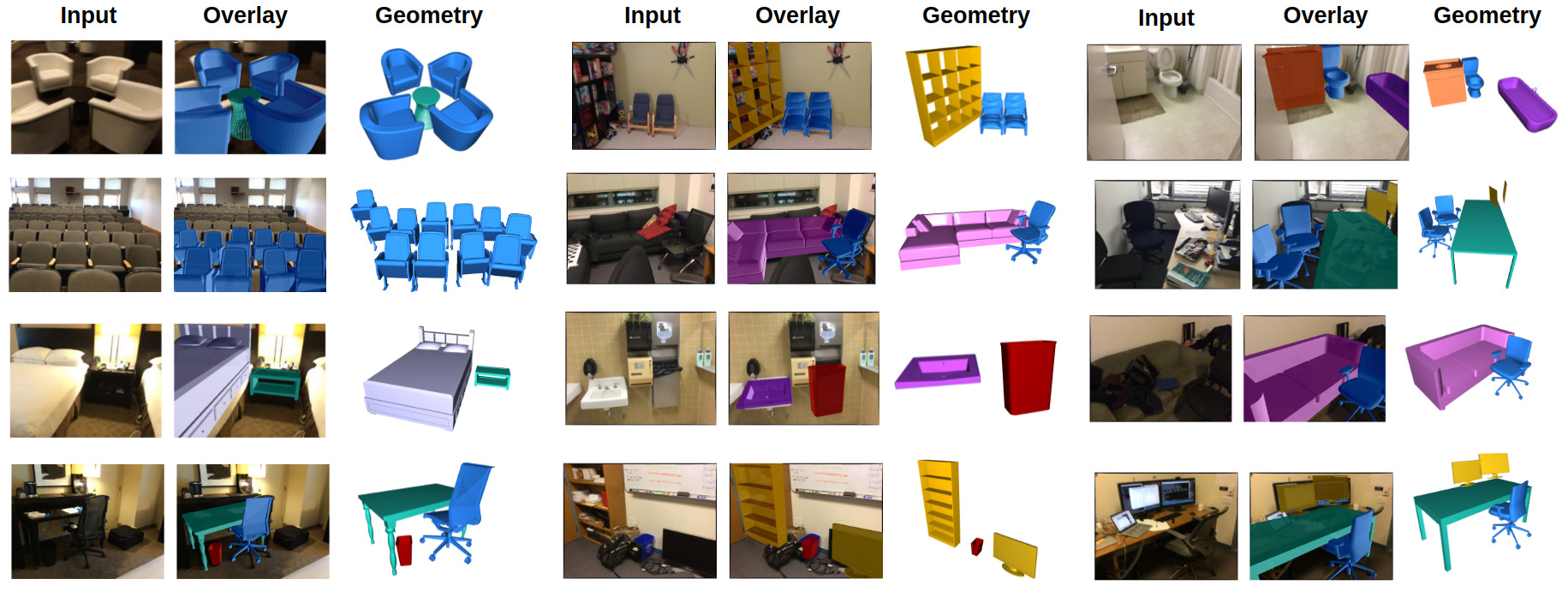}
    \vspace{-3.5mm}
    \caption{Sample predictions from our method. Our model shows promising 3D understanding ability in a wide variety of challenging real-world images \cite{scannet,scan2cad}.
    }
    \label{figure:sample_pred}
    \vspace{-0.5cm}
\end{figure*}

\subsection{Data and Evaluation}
\noindent\textbf{Dataset}. We evaluate our approach on the ScanNet25k image data \cite{scannet}, following state of the art for CAD alignment to images \cite{mask2cad, patch2cad, vid2cad}. 
This contains 20k training and 5k validation images, with training and validation images sampled from videos representing different scenes. 
We render Scan2CAD \cite{scan2cad} CAD annotations into the corresponding image views to obtain object detection, segmentation, depth, and NOC targets for training.
We consider input images at $360 \times 480$ resolution. 

\smallskip\noindent\textbf{Alignment Accuracy}. To evaluate our 9-DoF alignment performance, we adopt the alignment accuracy metric of Vid2CAD \cite{vid2cad}. 
For each scene, we first transform per-image 9-DoF alignment predictions to the ScanNet world space and apply a 3D clustering to predicted alignments, following the protocol described in \cite{vid2cad}. 
Alignment accuracy is then computed following  Scan2CAD \cite{scan2cad}: an alignment is correct if the corresponding object classification is correct, translation error $\leq 20$cm, rotation error is $\leq 20^\circ$, and scale ratio is $\leq 20\%$. 
Alignments can be predicted up to the total number of ground truth alignments.

\smallskip\noindent\textbf{Retrieval-Aware Alignment Accuracy}. The previous alignment accuracy metric, following \cite{vid2cad}, only takes into account object classification and 9-DoF alignment.
Thus, to evaluate retrieval, we define a retrieval-aware alignment accuracy metric. 
In addition to alignment correctness, this metric enforces the correctness of the retrieved CAD model.
Candidate CAD models are those that appear in the  ScanNet scene, following retrieval in prior works \cite{vid2cad, scan2cad, e2esocs}.

\subsection{Comparison to State of the Art}
We compare our method against the state-of-the-art 9-DoF Mask2CAD-b5 architecture~\cite{mask2cad, vid2cad}.
Since the original Mask2CAD predicted only 5-DoF alignments, Mask2CAD-b5 also learns to predict object depth and scale to produce 9-DoF alignments, as adapted by~\cite{vid2cad}.

We additionally compare our method against state-of-the-art single-image 3D object detectors, namely Total3D \cite{total3d} and MDR \cite{mdr}. Total3D~\cite{total3d} predicts from a single image the room layout, object poses and generates object meshes. To compare with their object pose estimation, we evaluate against their Object Detection Network (Total3D-ODN). MDR \cite{mdr} performs joint 3D object detection and voxel-based coarse-to-fine object reconstruction. We compare against their 3D CenterNet-based \cite{centernet} detector (MDR-CN). Since both methods depend on the room layout, we provide ground truth rotation. All baselines are trained on ScanNet25k data.
% since their object rotation estimation depends on room layout information that we do not have, we provide ground truth rotations to Total3D-ODN.

Table \ref{table:alignment} and Table \ref{table:retrieval_alignment} evaluate alignment accuracy and retrieval-aware alignment accuracy, respectively, in comparison with state of the art.
Our approach outperforms state of the art in alignment accuracy by 9.6\% (81\% relative) and by 9.5\% (53\% relative) in class and instance averages, due to our differentiable Procrustes-based alignment optimization. 
Additionally, when considering retrieval-aware alignment accuracy, \OURS{} improves by 8.1\% (85\% relative improvement) and by 7.9\% (57\% relative improvement) over state of the art in class and instance accuracies.
Our geometrically-grounded alignment formulation enables significant performance improvements in categories that are more scarcely represented in the train set (particularly ``bed", ``bathtub").

Figure~\ref{figure:qualitative_comp} shows a qualitative comparison of CAD retrieval and alignment on ScanNet images.
\OURS{} obtains more robust and accurate object alignments across a diverse set of image views and object types. 

\subsection{Ablations}
\noindent\textbf{Effect of end-to-end optimization on alignment}. Table \ref{table:alignment} shows that our end-to-end Procrustes optimization notably improves class alignment accuracy, from 14.9\% to 19.4\%.

\smallskip\noindent\textbf{Effect of learned Procrustes weights on alignment}. 
In Table~\ref{table:alignment}, adding learned weights to end-to-end Procrustes additionally improves robustness in alignment estimation, resulting in an improvement of average class alignment accuracy from 19.4\% to 20.4\%.

\smallskip\noindent\textbf{Effect of learned retrieval on alignment}.
In fact, introducing learned retrieval from the predicted NOC correspondences helps to improve class alignment accuracy from 20.4\% to 21.5\%, as shown in  Table \ref{table:retrieval_alignment}.
Here, the learned retrieval can provide additional signal to the correspondence learning, in our end-to-end formulation for joint retrieval and alignment.

\smallskip\noindent\textbf{Learned retrieval performance}.
In Table \ref{table:retrieval_alignment}, we compare our method with a baseline retrieval approach of using a single-sided Chamfer distance from the NOC predictions to the database models (referred as ``Ours P+E+W" in the tables). 
We achieve an improved performance when leveraging an end-to-end, learned approach that embeds NOC predictions into a shared space with CAD models.
\OURS{} additionally leverages a completion proxy loss for the retrieval embedding, producing the best retrieval-aware alignment performance.

\smallskip\noindent\textbf{Scaling to augmented training data}. 
\begin{table}[]
    \begin{center}
    \resizebox{0.32\textwidth}{!}{  
    \begin{tabular}{c| c c | c c}
        & \multicolumn{2}{c|}{Alignment} & \multicolumn{2}{c}{Retrieval} \\
        \hline
        Data & class & instance & class & instance \\
        \hline
        25k  & 21.5 & 27.4 & 17.6 & 21.7 \\
        400k & 24.4 & 31.9 & 20.2 & 24.9 \\
    \end{tabular}
    }
    \vspace{-3.5mm}
    \caption{Alignment and retrieval-aware alignment improvement due to augmenting the training image set, by sampling more frames from ScanNet \cite{scannet} videos. 
    }
    \label{table:400k}
    \vspace{-0.70cm}
    \end{center}
\end{table}
While the ScanNet25k benchmark contains 20k train and 5k validation images obtained by sampling every 100 frames from ScanNet videos \cite{scannet}. We consider augmenting training data by sampling images more densely. 
We prepare a 400k training dataset created by sampling every 5 frames, which helps to improve our alignment and retrieval-aware alignment performance, as shown in Table~\ref{table:400k}.
Although such dense frame sampling leads to stronger performance, such augmentation can only be obtained from video datasets. 
Thus, we focus on the 25k scenario, as it is more representative for image-based reconstruction.

\smallskip\noindent\textbf{Robustness analysis}.
\begin{table}
\begin{center}
\resizebox{0.32\textwidth}{!}{  
\begin{tabular}{c | c c c c}
 %\hline
 NOC noise $\sigma$ & 0.0 & 0.1 & 0.2 & 0.3 \\
 \hline
 Accuracy & 21.5 & 20.8 & 20.0 & 19.2 \\
 % \hline
\end{tabular}
}
\vspace{-3.5mm}
\caption{Ablation of robustness to NOC noise under simulated Gaussian noise.}
\label{tab:robust}
\vspace{-1.0cm}
\end{center}
\end{table}
We show in Table~\ref{tab:robust} that under simulated Gaussian noise in NOC predictions, our method maintains robustness, recovering 90\% of category average alignment accuracy even under strong noise of $\mathcal{N}(0, 0.3)$. 

%\paragraph{Limitations.} 
\medskip\noindent\textbf{Limitations.}
While \OURS{} shows significantly more robust and accurate CAD alignment performance to images, various limitations remain.
In particular, \OURS{} makes per-object predictions independently, while global scene context (e.g., object-object relationships) can provide additional important signal for object alignments.
Additionally, while CAD alignment and retrieval can provide an important semantic reconstruction of an observed scene, CAD databases cannot provide exact geometric matches in real-world scenarios, as is the case with Scan2CAD~\cite{scan2cad}. 
Thus, our approach is limited to approximate geometric representations; we believe that extending a CAD retrieval approach to couple with mesh deformation approaches (e.g., as in~\cite{ishimtsev2020caddeform}) is a promising direction to address more exact geometric reconstruction.

%% file: c5_conclusion.tex
\section{Conclusion}
\vspace{-0.1cm}
We have presented \OURS{}, a robust end-to-end approach for single-image CAD model alignment and retrieval. 
We show that leveraging dense per-pixel depth and canonical point correspondences with our weighted differentiable Procrustes optimization leads to more robust and accurate pose predictions.
Additionally, these correspondences can be leveraged for geometry-aware end-to-end retrieval to improve both retrieval and alignment performance.
For challenging ScanNet/Scan2CAD image data~\cite{scannet, scan2cad}, our method significantly improves state-of-the-art retrieval-aware alignment accuracy from 9.5\% to 17.6\%. Our approach runs efficiently at test time, achieving interactive speeds of 53 milliseconds per image. We hope that this can further spur developments in 3D perception towards content creation, mixed reality, and domain transfer scenarios.

%% file: c6_acknowledge.tex
% \section*{Acknowledgements}

\smallbreak\noindent\textbf{Acknowledgements.} This project is funded by the Bavarian State Ministry of Science and the Arts and coordinated by the Bavarian Research Institute for Digital Transformation (bidt), the TUM Institute of Advanced Studies (TUM-IAS), the ERC Starting Grant Scan2CAD (804724), and the German Research Foundation (DFG) Grant Making Machine Learning on Static and Dynamic 3D Data Practical. We also want to thank Kevis-Kokitsi Maninis and Vanessa Wirth for helping with the evaluation and Yinyue Nie for helping with the Total3D baseline.
%and useful discussions, 
% \newpage

%% file: c7_appendix.tex
\newpage
\section{Additional Results}
\subsection{Qualitative Evaluation on iPhone Images}
\begin{figure*}[!hb]
\begin{center}
    \includegraphics[width=\textwidth]{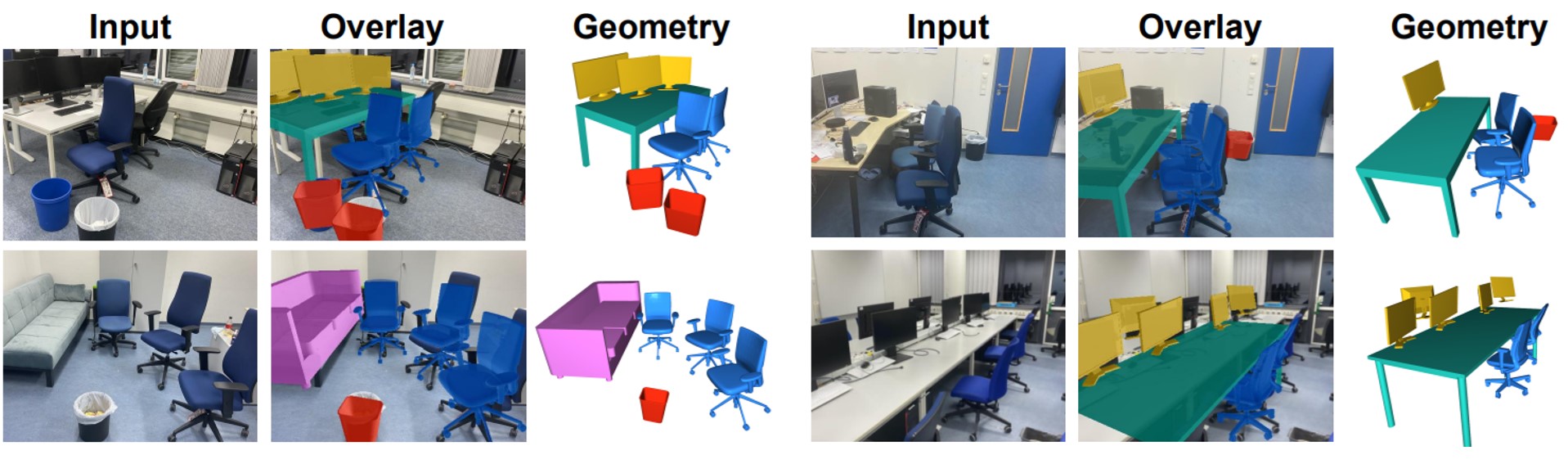}
    \vspace{-0.2cm}
    \caption{\textbf{Sample Alignments from Smartphone Images.} We take photos of a real work environment from an Iphone 11 rear camera. Our method achieves high quality alignments to complex scenes with multiple objects.}
    \label{figure:iphone}
\end{center}
\end{figure*}

We apply our approach trained on ScanNet+Scan2CAD to images taken with an iPhone 11 (rear camera) of various office environments.
As shown in Figure~\ref{figure:iphone}, our method achieves high-quality alignment results in complex scenes.

\subsection{Unconstrained Retrieval}
\begin{figure*}
    \centering
    \includegraphics[width=0.95\textwidth]{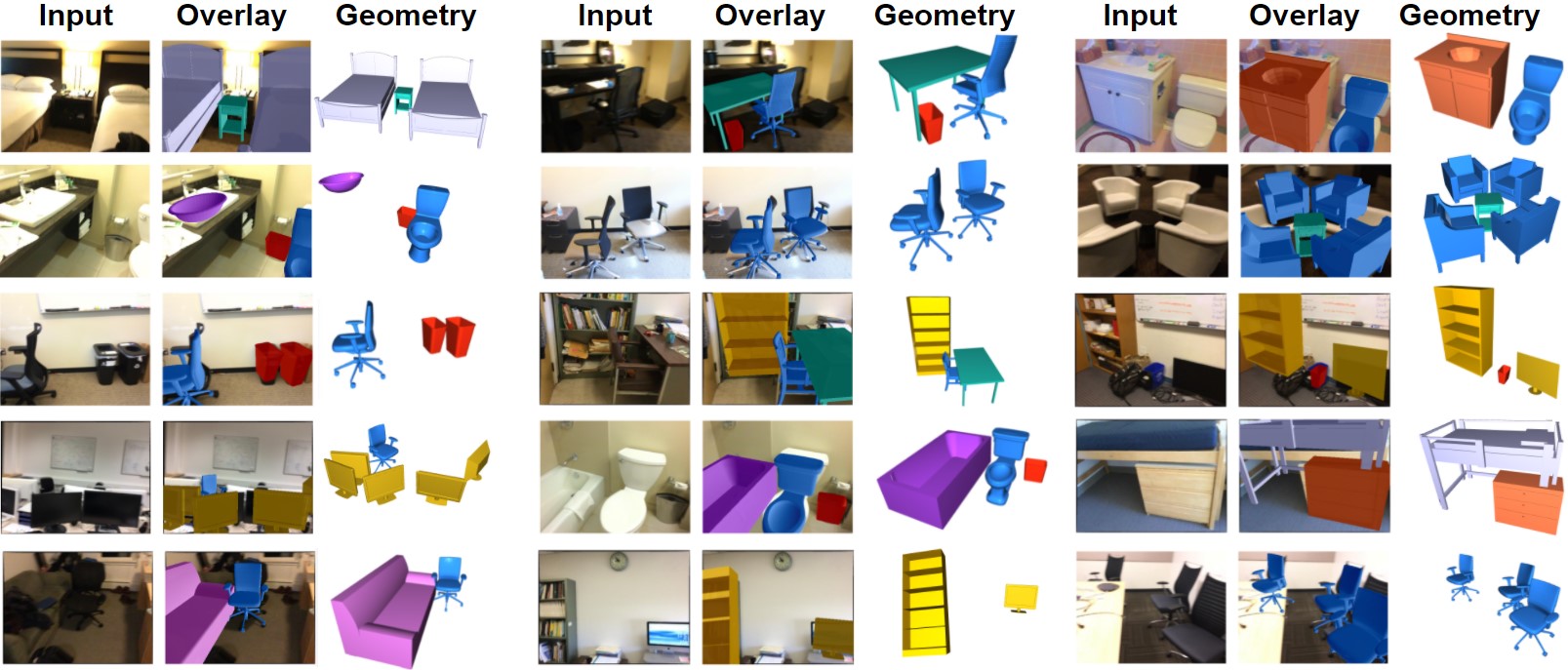}
    \vspace{-0.2cm}
    \caption{\textbf{Unconstrained Retrieval.} We show results from ScanNet validation set \cite{scannet}, where the candidate retrieval pool covers the full training set. Our method shows promising generalization capability in this challenging setup.}
    \label{figure:wild}
\end{figure*}

Following previous research \cite{scan2cad, e2esocs, vid2cad}, we retrieve CAD models from a given scene pool in our main results. In Figure~\ref{figure:wild}, we consider unconstrained retrieval where CAD models can be retrieved from the whole database of CAD models that appear in the training data. Note that the CAD models that only appear in the validation set are omitted. 

In the unconstrained setting, the number of unique CAD models from benchmark categories is $\approx 2300$. Similar to the standard approach, we pre-compute CAD embeddings prior to inference. Using per-category nearest-neighbor lookup from this large set, our method operates at 59 milliseconds per image at inference ($\approx 17$ frames per second). This is a 2-frame performance drop compared to the $\approx 19$ frames per second achieved with constrained CAD model pools. Even when retrieving from the large set of more than $2000$ CAD models, our method still achieves interactive frame rates and has the potential for real-time applications.

\section{Further Implementation Details}
\subsection{Depth Estimation Head}

\begin{figure}
\begin{center}
    \centering
    \includegraphics[width=.40\textwidth]{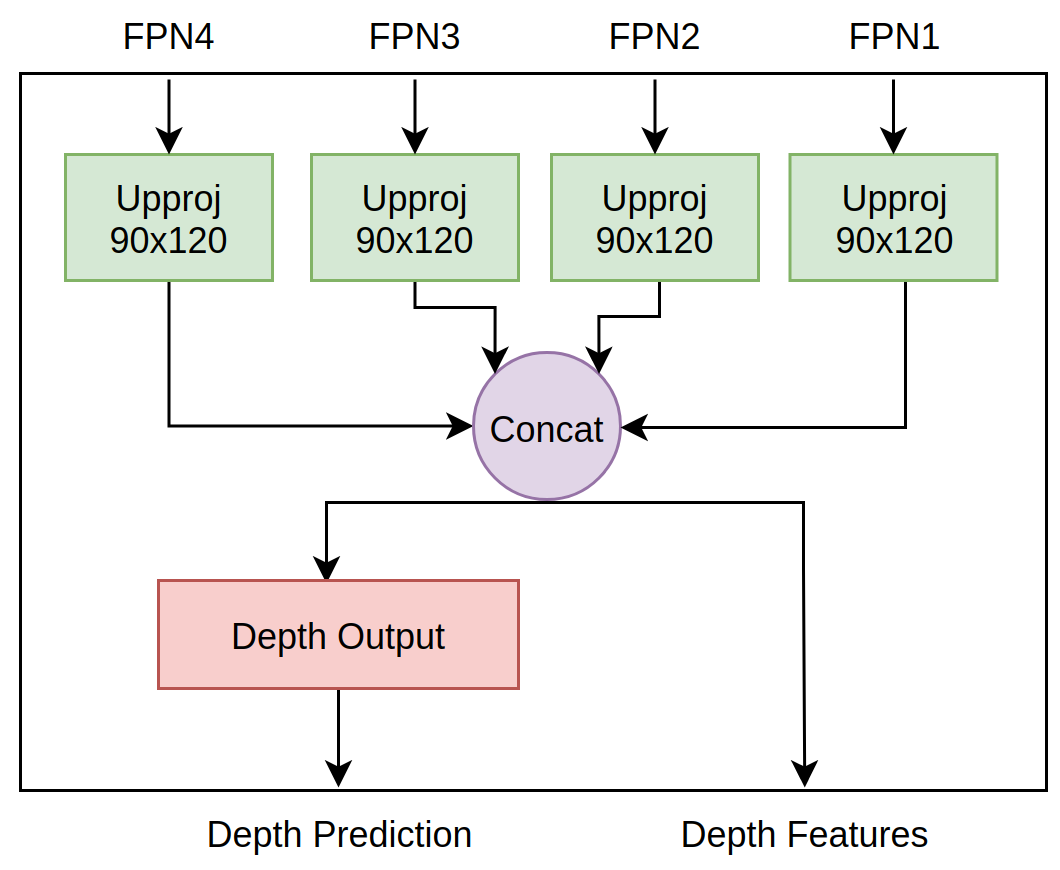}
    \caption{\textbf{Overview of Depth Prediction}. Adopted from \cite{ddp}, up-projection (Upproj) layers up-samples of to a given spatial resolution. The resulting multi-scale features are combined via concatenation and used in depth estimation and alignment pipeline.}
    \label{figure:depth_head}
\end{center}
\end{figure}

\begin{figure}
\begin{center}
    \centering
    \includegraphics[width=.48\textwidth]{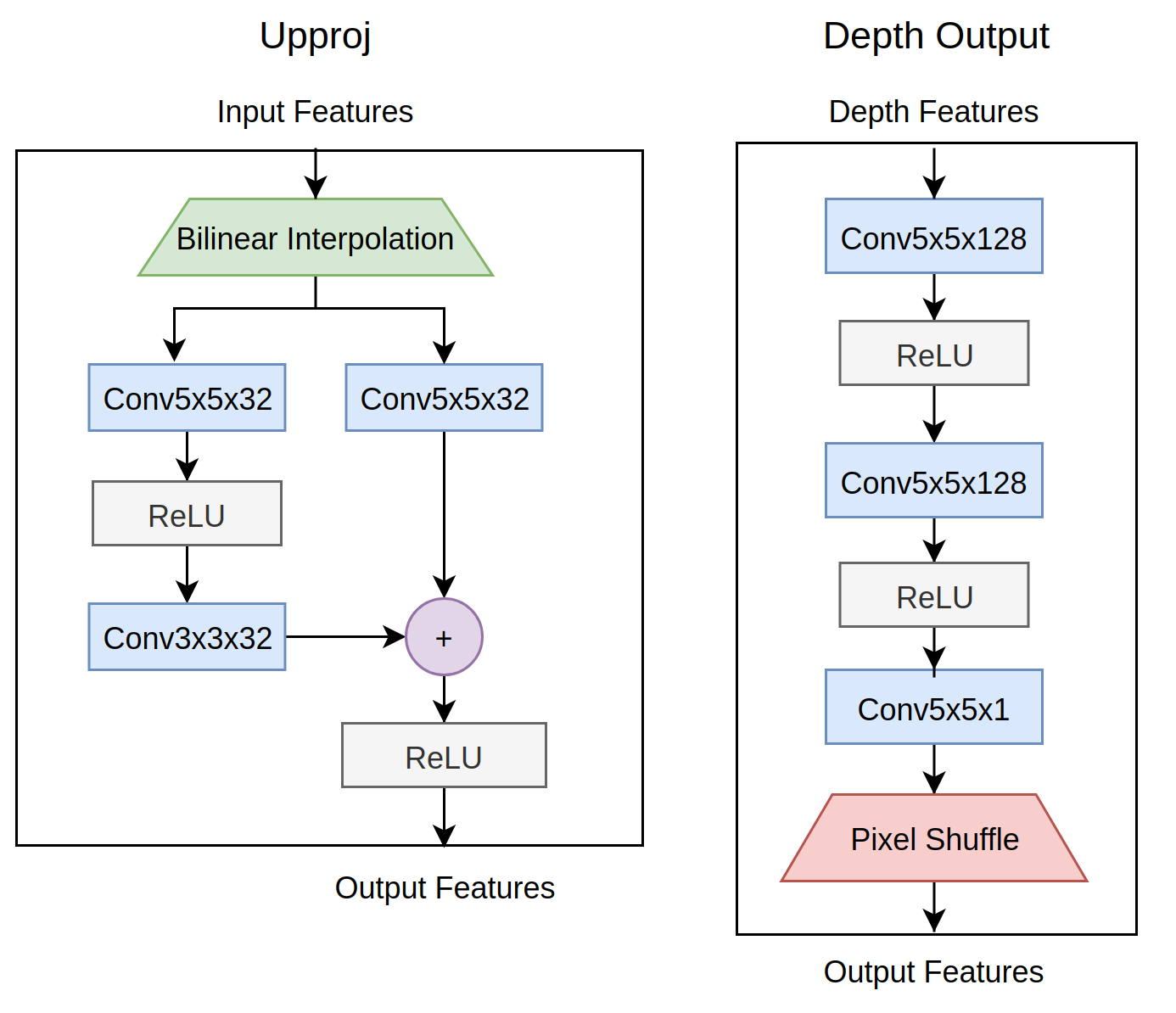}
    \caption{\textbf{Depth head modules}. Convolutions are described with their filter width, filter height, and output channels, respectively. Pixel shuffle layers up-sample spatial dimension by 4x4 (16 channels). The bilinear interpolation in Upproj layer adaptively rescales the input features to quarter of the image, 90x120 in out case, resolution irrespective of the input size. We use 32-channel feature maps at each up-projection following \cite{revisiting}.}
    \label{figure:depth_modules}
\end{center}
\end{figure}

To predict depth for an input image, we use the multi-scale future fusion (MFF) module from \cite{revisiting}. 
We adopt MFF to use FPN \cite{fpn} features instead of ResNet \cite{resnet} features directly, and omit the up-convolutional part for computational efficiency. Moreover, we use a pixel-shuffle \cite{superres} layer for a learnable, parametric up-sampling as opposed to the commonly used bilinear interpolation performed only in post-processing \cite{ddp, revisiting}.

Figure~\ref{figure:depth_head} diagrams our depth prediction head, and its sub-components are given in Figure~\ref{figure:depth_modules}.

To optimize depth estimation, we use the reverse Huber (berHu loss) \cite{ddp}, whose definition is
\begin{equation}
\textrm{berHu}(x) = \begin{cases} 
      |x| & |x| \leq c, \\
      \frac{x^2 + c^2}{2c} & |x| > c, \\
   \end{cases}
\end{equation}
where $c$ is set adaptively to $\frac{1}{5}$th of the loss values in a grid. This loss allows low-error predictions to get more accurate by down-weighting the outliers during training.

\subsection{Retrieval Network}
\begin{figure*}[h]
    \centering
    \includegraphics[width=\textwidth]{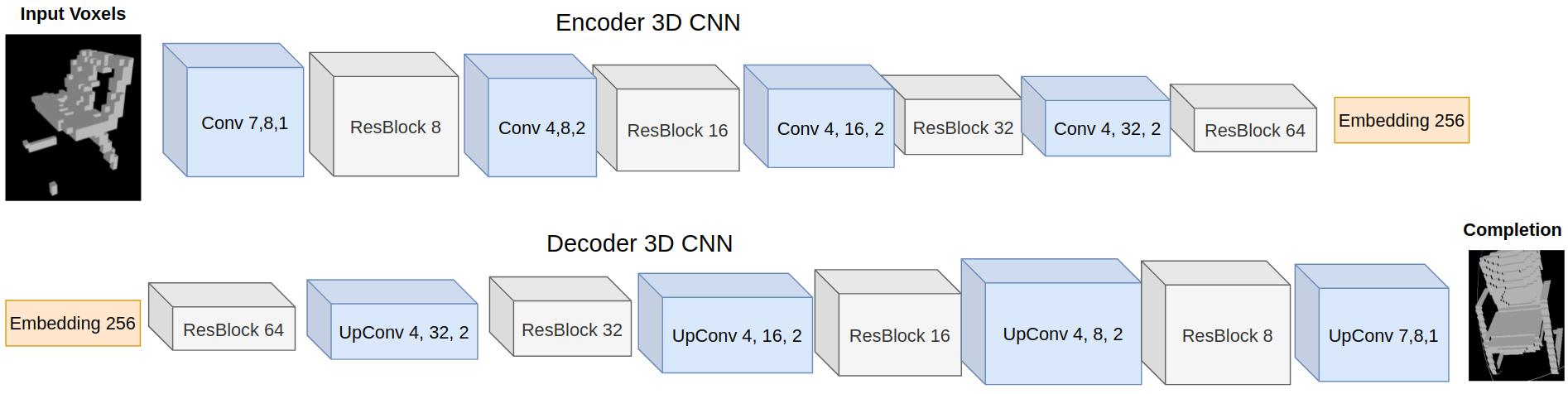}
    \caption{\textbf{NOC-based Retrieval of CADs to Images.} 
    We use our predicted NOCs to enable 3D-based joint embedding of real-world observations and CAD models, by interpreting the NOCs as a voxelization in the canonical space. 
    Conv and UpConv represent convolution and transpose convolution layers, with comma-separated values represent filter size, output channels, and stride, respectively. Each Conv and UpConv layer are followed by a ReLU activation. ResBlocks represent ResNet basic blocks \cite{resnet} that use 3D convolutions. The result of the encoder and input for the decoder is a 256-dimensional embedding vector.}
    \label{figure:ret_modules}
\end{figure*}

We show the retrieval part of our architecture in Figure~\ref{figure:ret_modules}, which is inspired by the joint scan-CAD embedding of \cite{jointemb}.

\subsection{Data Preparation}
We render Scan2CAD alignment labels \cite{scan2cad} over ScanNet images \cite{scannet} using a simple rasterization pipeline \cite{raster}. Before rendering, alignment labels are projected to the image camera coordinate systems, using the inverse of the camera pose provided by the ScanNet \cite{scannet}.

Following Mask2CAD \cite{mask2cad}, we filter out objects whose centers are outside the image frame. Furthermore, no ScanNet labels except for Scan2CAD alignments, camera poses and camera intrinsics are involved in the rendering pipeline. Thus, the rendering and supervision of our method is consistent with and directly comparable to the previous work \cite{mask2cad, vid2cad}.

In the dense frame sampling experiment, we extracted images from ScanNet raw sensor stream and perform the same data processing.

\subsection{Total3D Training Details}
We re-train the pose estimation component of Total3D \cite{total3d} on our data for comparison. Since the method relies on pre-computed object detections, we first train a ResNet50 Faster-rcnn \cite{fpn, detectron2} initialized from ImageNet and COCO pretraining. Then, we match the pre-computed detections with our labels based on bounding box IOUs. Also using the pre-computed detections, we extract the relevant geometry features for the pose estimation pipeline \cite{total3d}.

Since the rotation component of total3d depends on the room layout that we do not have, we simply use the ground-truth rotation. 

We use an SGD optimizer similar to our main training. We train for the total of 60k iterations, decaying learning rate at 40k, based on our tuning of the model.

\subsection{Used Open-Source Libraries}
We utilize various open source libraries for our model trainings and data pre-processing. Our model is implemented using PyTorch, Detectron2, and Pytorch3D \cite{pytorch, detectron2, pytorch3d}, without declaring any custom low-level kernels outside of these libraries. We make our geometry visualizations using Open3D \cite{open3d} and CAD voxelizations using Trimesh \cite{trimesh}. For the baseline methods without learned retrieval, we performed single sided Chamfer Distance lookup over point clouds sampled using the farthest points sampling implementation from PyTorch Cluster \cite{pg}.